\documentclass[10pt,conference]{IEEEtran}
\IEEEoverridecommandlockouts

\makeatletter
\newcommand{\DeclareLatinAbbrev}[2]{%
  \DeclareRobustCommand{#1}{%
    \@ifnextchar{.}{\textit{#2}}{%
      \@ifnextchar{,}{\textit{#2.}}{%
        \@ifnextchar{!}{\textit{#2.}}{%
          \@ifnextchar{?}{\textit{#2.}}{%
            \@ifnextchar{)}{\textit{#2.}}{%
              {\textit{#2.,\ }}}}}}}}%
}
\makeatother
\DeclareLatinAbbrev{\eg}{e.g}
\DeclareLatinAbbrev{\Eg}{E.g}
\DeclareLatinAbbrev{\ie}{i.e}
\DeclareLatinAbbrev{\Ie}{I.e}
\DeclareLatinAbbrev{\etc}{etc}
\DeclareLatinAbbrev{\etal}{et~al}

\def\first {$(i)$\xspace}
\def\Second{$(ii)$\xspace}

\def\third {$(iii)$\xspace}
\def\fourth{$(iv)$\xspace}
\def\fifth {$(v)$\xspace}

\newcommand{\smartparagraph}[1]{\vspace{.05in}\noindent\textbf{#1}}

\DeclareRobustCommand{\circled}[1]{\tikz[baseline=(char.base)]{
            \node[shape=circle,fill=black,inner sep=0pt, minimum size=12pt,scale=0.9] (char) {\textcolor{white}{#1}};}}

\usepackage[utf8]{inputenc} 
\usepackage[T1]{fontenc}    
\usepackage[colorlinks=true, allcolors=black]{hyperref}
\usepackage{url}            
\usepackage{booktabs}       
\usepackage{amsfonts}       
\usepackage{nicefrac}       
\usepackage{microtype}      
\usepackage{xcolor}         

\usepackage{cite}
\usepackage{graphicx}
\usepackage{float}
\usepackage{xspace}
\usepackage{tabularx}
\usepackage{longtable} 
\usepackage{colortbl}
\usepackage{siunitx}
\usepackage{multirow}
\usepackage{makecell}
\usepackage{dcolumn}
\usepackage{multirow}
\usepackage{pifont}
\usepackage{placeins}
\usepackage{tikz}
\usepackage{amsmath}

\makeatletter
    \newcommand{\linebreakand}{%
      \end{@IEEEauthorhalign}
      \hfill\mbox{}\par
      \mbox{}\hfill\begin{@IEEEauthorhalign}
    }
\makeatother

\makeatletter
\let\MYcaption\@makecaption
\makeatother

\usepackage[font=footnotesize]{subcaption}

\makeatletter
\let\@makecaption\MYcaption
\makeatother

\newcolumntype{d}[1]{D{.}{.}{#1}}
\def\BibTeX{{\rm B\kern-.05em{\sc i\kern-.025em b}\kern-.08em
    T\kern-.1667em\lower.7ex\hbox{E}\kern-.125emX}}

\graphicspath{{Images/}} 

\title{Deriving Coding-Specific Sub-Models \\from LLMs using Resource-Efficient Pruning}

\author{\IEEEauthorblockN{Laura Puccioni$^{\dagger}$}
\IEEEauthorblockA{\textit{Spotify}\thanks{
$^{\dagger}$Work performed while at RISE AB.}}
\and
\IEEEauthorblockN{Alireza Farshin}
\IEEEauthorblockA{\textit{NVIDIA}}
\and
\IEEEauthorblockN{Mariano Scazzariello}
\IEEEauthorblockA{\textit{RISE AB}}
\linebreakand
\IEEEauthorblockN{Changjie Wang}
\IEEEauthorblockA{\textit{KTH Royal Institute of Technology}}
\and
\IEEEauthorblockN{Marco Chiesa}
\IEEEauthorblockA{\textit{KTH Royal Institute of Technology}}
\and
\IEEEauthorblockN{Dejan Kosti\'c}
\IEEEauthorblockA{\textit{KTH Royal Institute of Technology} \\ \textit{RISE AB}}
}

\begin{document}
\bstctlcite{IEEEexample:BSTcontrol}

\maketitle

\begin{abstract}
Large Language Models (LLMs) have demonstrated their exceptional performance in various complex code generation tasks. However, their broader adoption is limited by significant computational demands and high resource requirements, particularly memory and processing power. To mitigate such requirements, model pruning techniques are used to create more compact models with significantly fewer parameters. However, current approaches do not focus on the efficient extraction of programming-language-specific sub-models. In this work, we explore the idea of efficiently deriving coding-specific sub-models through unstructured pruning (\ie Wanda). We investigate the impact of different domain-specific calibration datasets on pruning outcomes across three distinct domains and extend our analysis to extracting four language-specific sub-models: Python, Java, C++, and JavaScript. We are the first to efficiently extract programming-language-specific sub-models using appropriate calibration datasets while maintaining acceptable accuracy w.r.t. full models. We are also the first to provide analytical evidence that domain-specific tasks activate distinct regions within LLMs, supporting the creation of specialized sub-models through unstructured pruning. We believe that this work has significant potential to enhance LLM accessibility for coding by reducing computational requirements to enable local execution on consumer-grade hardware, and supporting faster inference times critical for real-time development feedback.
\end{abstract}


\section{Introduction}
\label{sec:introduction}
Large Language Models (LLMs) have become the forefront of technological advancements, making significant strides in various fields thanks to their exceptional performance across diverse and complex tasks. Recently, LLMs have been extensively used in code generation~\cite{10.1145/3643795.3648377,google-ai-code}, either acting as co-pilots (\ie assisting developers)~\cite{gh-copilot,tabby-copilot} or even automating parts of the development process~\cite{cursor,devin}. However, despite their growing use in code-related tasks, their broader adoption is limited by significant computational demands and high resource requirements of forefront models, especially in terms of memory and processing power~\cite{back_5}. Over the past year, model sizes have continuously increased, with Llama-3.1, the largest open-source model, reaching 405 billion parameters~\cite{llama3.1}. As the size of these models continues to grow, with GPT-5 allegedly exceeding 10 trillion parameters~\cite{gpt5}, the challenge of efficiently inferencing becomes significant, especially as LLMs deployment poses an ever-increasing energy demand that may soon become unsustainable~\cite{ai-exhausting}.

\smartparagraph{One model does not fit all.} General-purpose LLMs are trained on vast datasets spanning multiple domains, with their parameter count carefully set to optimize the balance in the loss function, largely influenced by the volume of training data. However, certain applications (\eg coding or math) do not require broad knowledge but instead benefit from expertise in a specific domain. For example, developers might find a model specialized in a particular programming language, such as Python, more helpful for coding tasks, without needing knowledge on unrelated subjects like physics or chemistry. In such instances, a domain-specific sub-model is ideal, as it can maintain essential human language understanding along with specialized domain expertise.

\smartparagraph{Efficiency via compression.} In this context, model compression techniques have proven effective. These approaches aim to produce a compact\slash sparse model with significantly fewer parameters, enhancing deployment efficiency in terms of both cost and resource usage.
One prominent approach in this domain is pruning, which involves removing unnecessary parameters from a model without significantly degrading its performance~\cite{prune}. 
However, many existing pruning methods~\cite{NIPS2015_ae0eb3ee,xia-etal-2022-structured} are not specifically devised for LLMs, causing significant inaccuracies, and often require retraining, making them less feasible for models of such scale and complexity. Such limitations have prompted the need for more tailored approaches that can address the specific challenges posed by these models.
Wanda~\cite{wanda} presents a novel technique for pruning LLMs by leveraging a small set of calibration data to estimate activation norms. This method efficiently identify less critical weights\slash parameters in the model, allowing for pruning with lower computational\slash memory demands and minimal accuracy drops of the pruned LLM. Importantly, Wanda achieves this without the need to retrain or fine-tune the full model.
Although Wanda offers several advantages, it is primarily designed to extract general-purpose sub-models that maintain strong performance across multiple tasks, rather than focusing on domain-specific sub-models. Furthermore, authors do not investigate the impact of calibration sample selection on pruning outcomes, relying exclusively on samples from the C4 dataset~\cite{2019t5}, a broad English-language text source.

\smartparagraph{Scope of the paper and contributions.} In this paper, we aim to explore whether resource-efficient pruning techniques (\ie Wanda) can be successfully applied to extract domain-specific sub-models without requiring retraining. We also investigate how the selection of various domain-specific calibration samples impacts pruning results. 
Our primary focus is on sub-model extraction for code generation, specifically Python, Java, C++, and JavaScript. To further validate our approach, we also extract sub-models in the domains of math, common-sense reasoning (CSR), and language translation.
Our results demonstrate that, to the best of our knowledge, we are the first ones to successfully extract programming-language-specific sub-models from foundational models using appropriate calibration sets, and that such sub-models retain acceptable accuracy w.r.t. full models. Furthermore, we validate that sub-models extracted using domain-specific datasets outperform those derived from unrelated domain datasets in specialized tasks (\eg code generation). Lastly, we provide the first analytical evidence that domain-specific tasks activate different regions within LLMs when applying resource-efficient unstructured pruning.   
%

\smartparagraph{Impact.} We believe that domain-specific sub-models offer several impactful advantages for coding-related tasks: \first compact models reduce computational demands, making it feasible to run them locally on consumer-grade hardware, which broadens access to LLMs and democratizes their use for developers with limited resources; \Second by employing local, domain-specific sub-models for critical domains (\eg military defense systems or cybersecurity), companies can ensure that proprietary or confidential data remains in-house, minimizing reliance on external APIs and enhancing data security; \third sub-models enable faster inference times, which is particularly valuable for tasks requiring real-time feedback, such as iterative coding\slash debugging; \fourth our approach supports an always-updated collection of specialized models that can be maintained independently, simplifying the rollout of improvements or domain-specific updates without compromising the accuracy of other models; and \fifth our method allows individual sub-models to be fine-tuned or adjusted without altering an entire general-purpose LLM, offering flexibility to create highly customized sub-models that closely meet project-specific needs.
\section{Related Work}\label{sec:background}

LLMs are decoder-only models, meaning they only utilize the decoder part of the Transformer architecture~\cite{attention}. In such models, input tokens are transformed into embeddings that are directly fed into the decoder, which processes them through a series of matrix multiplications involving a large number of parameters. Given the high computational demand of these multiplications, significant research efforts have focused on model compression, which aims to transform large, resource-intensive models into more compact versions~\cite{compr}. These smaller models are suitable for deployment on devices with less sophisticated hardware and are optimized for faster execution with minimal latency~\cite{back_7}.

\smartparagraph{Model compression approaches.} In recent research efforts, five primary model compression approaches have emerged: \first quantization, \Second knowledge distillation, \third Neural Architecture Search (NAS), \fourth low-rank factorization, and \fifth pruning. Each of these methods offers unique advantages and challenges, and are complementary. Quantization is a compression method that reduces model size by representing weights and activations with lower-bit representations, decreasing computational complexity and storage requirements~\cite{compr_tech}. The most common approach is post-training quantization, which converts pretrained weights to lower bit precision~\cite{quant_6}. However, quantization may introduce errors that impact the model's ability to generalize or maintain performance across varied tasks, lowering overall accuracy. Low-rank factorization compresses models by decomposing weight matrices into lower-rank matrices, minimizing redundancy and facilitating faster computations by parallelizing memory access of dense matrices~\cite{lowrank_application, lowrank_application2}. Knowledge distillation transfers knowledge from a large model (teacher) to a smaller model (student), aiming to replicate the teacher's behavior with fewer resources~\cite{back_knowdist,dist_3, dist_4}. NAS is similar to knowledge distillation and often involves training a super-network once, enabling the sampling of sub-networks through the weight-sharing principle~\cite{nas,llamanas}. Low-rank factorization, knowledge distillation, and NAS all involve retraining or significant modifications to the model architecture, making them challenging to implement for LLMs.
Among these approaches, pruning has proven to offer the best trade-off between computational gains and minimal loss in accuracy. It aims to reduce the size or complexity of a model systematically, eliminating redundant or less influential parameters, thereby reducing the model's computational and storage requirements~\cite{pruning_1}. In this work, we mainly focus on unstructured pruning, \ie induce sparsity at a finer granularity by eliminating individual weights rather than entire units or blocks (\ie neurons or layers)~\cite{pruning_2}.

\begin{table}[t]
\centering
\caption{Qualitative comparison among pruning approaches.}
\resizebox{\linewidth}{!}{%
\begin{tabular}{c c c c c} 
\textbf{} & \textbf{No Retraining?} & \textbf{Efficient?} & \textbf{Domain Specific?} & \textbf{w/ Calib. Dataset}\\
 \hline
  Magnitude~\cite{pruning_1} & {\color{red}\ding{55}} & {\color{green}\ding{51}} & {\color{red}\ding{55}} & {\color{red}\ding{55}}\\
  MaskLLM~\cite{maskllm} & {\color{red}\ding{55}} & {\color{green}\ding{51}} & {\color{red}\ding{55}} & {\color{red}\ding{55}}\\
 SparseGPT~\cite{SparseGPT} & {\color{green}\ding{51}} & {\color{red}\ding{55}} & {\color{red}\ding{55}} & {\color{green}\ding{51}}\\
 Wanda~\cite{wanda} & {\color{green}\ding{51}}  & {\color{green}\ding{51}} &  {\color{red}\ding{55}} & {\color{green}\ding{51}}\\
 D-Pruner~\cite{dpruner} & {\color{red}\ding{55}} & {\color{red}\ding{55}} & {\color{green}\ding{51}} & {\color{green}\ding{51}}
\end{tabular}
}
\label{table:pruning-comparison}
\vspace{-.2in}
\end{table}

\smartparagraph{Existing pruning approaches.} Several existing works, summarized in Table~\ref{table:pruning-comparison}, propose various methods for pruning. Magnitude pruning~\cite{pruning_1} removes low-magnitude weights, assuming these contribute minimally to the model's function, thus improving inference speed. While computationally efficient, it requires network fine-tuning and may prematurely remove weights that may become significant as pruning progresses~\cite{magn}. This technique works well for smaller models, but it dramatically fails for LLMs even with relatively low levels of sparsity.
MaskLLM~\cite{maskllm} introduces an approach for pruning LLMs by incorporating mask learning directly into the training process. Unlike post-training pruning, MaskLLM learns structured sparsity patterns during model training, enabling the network to identify and retain essential weights while discarding redundant ones. While MaskLLM effectively reduces model size and computational demands, it requires modifications to the training pipeline.
SparseGPT~\cite{SparseGPT} uses a one-shot, layer-wise pruning strategy that avoids retraining by tackling a layer-wise reconstruction problem. This approach compresses each layer independently, aiming for sufficient accuracy by minimizing the error between original and pruned layers. Despite SparseGPT avoids retraining, it requires matrix inversion operations, which, while contributing to its effectiveness, can pose computational challenges~\cite{back_9}.
Wanda~\cite{wanda}, which stands for ``\textbf{W}eights \textbf{and a}ctivations'' is one of the most innovative pruning methods designed for LLMs, balancing computational simplicity with the advantage of avoiding retraining. Wanda evaluates the significance of each weight by computing the product of its magnitude and the norm of the corresponding input activations, estimated using a small set of calibration data. It then ``zeros-out'' the weights with the lowest products within each layer. While Wanda is effective for creating general-purpose sub-models, authors only rely on the C4 dataset~\cite{2019t5} for calibration, without exploring how different calibration samples might influence pruning results.
D-Pruner~\cite{dpruner} is a novel approach for creating domain-specific LLMs from a foundational model, leveraging full-parameter fine-tuning and gradient updates. Although effective, D-Pruner is memory\nobreakdash-intensive due to the full-parameter fine-tuning and has higher computational complexity because of the gradient updates. In contrast, Wanda is more efficient, as it avoids gradient updates, weight reconstruction, and retraining, requiring only a single forward pass.
\section{Methodology}\label{sec:proposed_approach}

Prior works examined how calibration data affect the effectiveness of model compression methods~\cite{back_9,calibr_1,calibr_2, calibr_3}, offering two key insights: \first no single dataset consistently performs best across all tasks, emphasizing the sensitivity of compression outcomes to the choice of calibration data, and \Second the optimal size for calibration sets is relatively small, as larger datasets yield minimal performance improvements, suggesting that a carefully selected, compact set is sufficient for effective pruning.
D\nobreakdash-Pruner~\cite{dpruner} is the first work to use task-specific calibration sets to extract domain-specific sub-models, focusing primarily on financial and medical domains, but not addressing programming or code generation tasks. Furthermore, as discussed in \S\ref{sec:background}, D-Pruner relies on full-parameter fine-tuning and gradient updates, which results in high computational demands.

\begin{figure}[t]
    \centering
    \includegraphics[width=\linewidth]{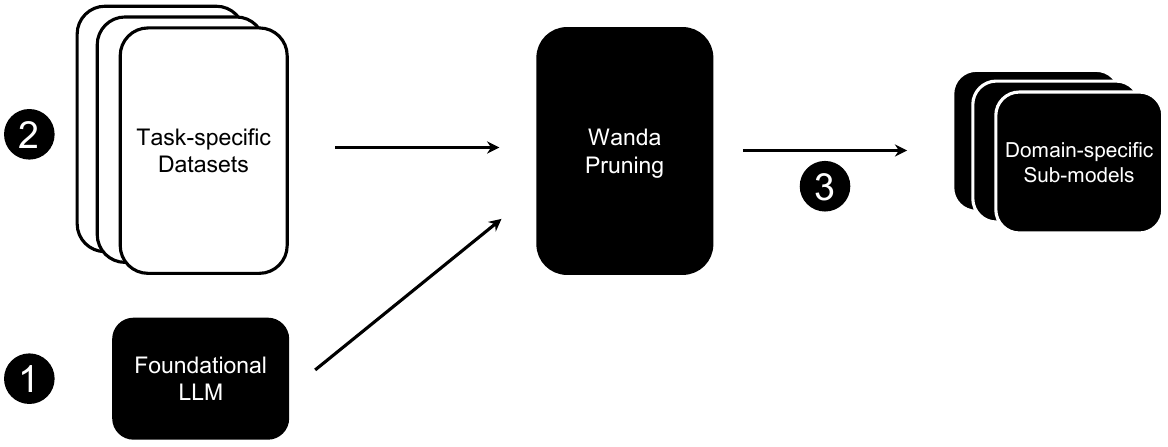}
    \caption{Our methodology for domain-specific-LLMs extraction.}
    \label{fig:system}
    \vspace{-.2in}
\end{figure}

Following the approach shown in Fig.~\ref{fig:system}, we assess whether combining efficient pruning techniques (\ie Wanda) with task-specific calibration datasets can effectively generate domain-specific LLMs that maintain strong performance in specialized tasks while minimizing memory and computational demands. We first validate our approach by extracting sub-models in the domains of math, CSR, and language translation. Then, we focus on sub-model extraction for code generation, \ie Python, Java, C++, and JavaScript. 

\vspace{.05in}
\noindent Our methodology consists of three main steps, which will be detailed in the following.

\smartparagraph{\circled{1} Model selection.} We begin by selecting one or more models for pruning. To ensure the study's validity and broad applicability, we chose a diverse range of models based on factors such as novelty, size, and purpose. 
First, we selected the latest versions of models (available at the time of writing) to enable fair and relevant comparisons. Second, our focus is on relatively compact models, as our goal is to create domain-specific sub-models that can be efficiently deployed on commodity hardware with limited resources. Finally, models were chosen based on their performance on widely recognized evaluation tasks, ensuring a trustable baseline for fair comparison between the original model and its pruned counterpart. While the original Wanda paper focused solely on Llama-2~\cite{back_9}, this study includes four diverse models, specifically: CodeLlama-7B and CodeLlama-13B~\cite{codellama}, Mistral-7B-v0.1~\cite{mistral7b}, and Gemma-1.1-7B~\cite{gemma}. CodeLlama is fine-tuned specifically for code, while Mistral and Gemma are more general-purpose models. We include two CodeLlama versions to validate our findings on a larger model. All the LLMs were chosen in their ``instruct'' version, as it is designed to follow natural language instructions more seamlessly.

\begin{table}
\centering
\caption{Selected datasets for pruning.}
\vspace{-.2in}
\begin{subtable}{.6\linewidth}
\caption{}
{
    \centering
    \begin{tabular}{c c} 
    \textbf{Domain} & \textbf{Datasets}\\
     \hline
      Code Generation & \cite{python_alpaca} \cite{github_python} \cite{OpenBookQA2018}\\
      Math & \cite{amini-etal-2019-mathqa} \cite{yue2023mammoth} \cite{mitra2024orcamath}\\
      CSR & \cite{OpenBookQA2018} \cite{zellers2019hellaswag} \cite{talmor-etal-2019-commonsenseqa}\\
      Translation & \cite{wmt} \cite{song_transl} \cite{tiedemann-2012-parallel}
    \end{tabular}
}
\label{table:domain-specific-datasets}
\end{subtable}
\begin{subtable}{.3\linewidth}
\caption{}
{
    \centering
    \begin{tabular}{c c} 
    \textbf{Language} & \textbf{Dataset}\\
    \hline
    Python & \cite{python_alpaca}\\
    Java & \cite{husain2019codesearchnet}\\
    C++ & \cite{nhlcoding_cleaned_cpp_dataset}\\
    JavaScript & \cite{husain2019codesearchnet}
    \end{tabular}
}
\label{table:programming-languages-datasets}
\end{subtable}
\vspace{-.25in}
\end{table}

\smartparagraph{\circled{2} Datasets selection.} As mentioned in \S\ref{sec:background},  Wanda utilizes a calibration set to estimate input activations essential for identifying weights to prune, making the quality and selection of these datasets critical for obtaining sub-models with high accuracy. For code generation, we employed: \first three Python-specific datasets to validate the approach (see Table~\ref{table:domain-specific-datasets}), and \Second four datasets covering different programming languages to generalize our findings to other coding tasks, reported in Table~\ref{table:programming-languages-datasets}. As for math, CSR, and language translation tasks, we chose the three datasets listed in Table~\ref{table:domain-specific-datasets}.

\smartparagraph{\circled{3} Pruning process.} We apply the Wanda pruning technique to the selected models and calibration datasets. Following prior best practices~\cite{back_9}, we use 128 samples as calibration data for each dataset considered. Samples are randomly selected with a fixed seed. Given the diverse datasets and models presented in Step~\circled{2}, the pruning process is performed iteratively. This iterative approach systematically adjusts parameters related to the model, sparsity type, and datasets, allowing to assess how different configurations affect pruning results. Although our experiments primarily focuses on 50\% unstructured sparsity, our takeaways are also applicable to structured pruning (\eg structured N:M sparsity~\cite{2_4_sparsity}), as shown by Wanda~\cite{wanda}. 
After completing the pruning process, we obtain a total of 4 programming-language-specific sub-models for each base model (\ie from Table~\ref{table:programming-languages-datasets}) and 13 sub-models for each base model: 12 task-specific sub-models obtained using the datasets in Table~\ref{table:domain-specific-datasets} and one additional sub-model using the C4 dataset to compare against generic (non-domain-specific) pruning.

\smartparagraph{Sub-models accuracy evaluation.} The accuracy of the obtained sub-models is then evaluated using different types of benchmarks and metrics tailored to the considered domain. For the code generation task, we evaluate performance using the \textit{pass@k} metric~\cite{chen2021evaluating} on two standard benchmarks~\cite{chen2021evaluating,mbpp}. The \textit{pass@k} metric is chosen for its ability to assess the functional correctness of generated code, rather than only checking for semantic similarity to the reference solution. Specifically, we set $k=10$ in our experiments. For CSR, following prior studies, we use a straightforward \textit{zero-shot evaluation} across six datasets~\cite{OpenBookQA2018,zellers2019hellaswag,ai2:winogrande,clark2019boolq,arc}. For the language translation task, we selected the \textit{METEOR} metric~\cite{meteor} and evaluated sub-models on two French-English datasets~\cite{wmt,tiedemann-2012-parallel}. \textit{METEOR} matches unigrams between machine-generated and human translations. For each test sample, we generate five translations, compare them to the reference, and retain the highest \textit{METEOR} score among the five outputs. For the math task, we selected the GSM8K dataset~\cite{cobbe2021gsm8k}. To align with evaluation methods in other studies, we calculate accuracy using \textit{8-shot accuracy}, where the model is given eight examples, each with a problem, solution, and a brief reasoning explaining the solution.

\vspace{.05in}
It is worth noticing that, although some evaluation datasets are also used as calibration datasets, we use separate splits to prevent overlap. In this way, we minimize the risk of overfitting, preserving the validity of the results.


\smartparagraph{Sub-models comparison.} Our goal is to investigate how various calibration datasets influence pruning outcomes and to determine whether pruning techniques can effectively create domain-specific sub-models. To validate our hypothesis, we conduct a fine-grained comparison of the parameter masks of each layer of the pruned models. We first use the \textit{penzai} tool~\cite{penzai} to visually represent the weight matrices of pruned sub-models. This visualization method allows us to clearly depict which weights were preserved and which were discarded across sub-models after pruning. By overlaying these matrices from sub-models belonging to the different domains, we aim to identify significant differences that could highlight the impact of the calibration datasets. Following a qualitative analysis, we validate our findings using a per-layer Jaccard distance measurement~\cite{jaccard_img}. This metric allows us to quantify the similarity of retained weights across layers, allowing us to quantitatively evaluate the differences between sub-models within the same domain and those across different domains, highlighting differences based on domain specificity.

\section{Evaluation and Discussion}

In this section, we discuss the results of the domain-specific sub-models created by pruning the foundational models described in \S\ref{sec:proposed_approach} with the chosen calibration datasets. We begin by reviewing the performance of sub-models generated for the four selected domains, followed by an analysis of the programming-language-specific sub-models (\S\ref{ss:task-specific}). Additionally, we dig into the internal structure of the obtained sub-models using both visual and analytical metrics (\S\ref{ss:submodel-structure}).

\subsection{Task-Specific Pruning Evaluation}\label{ss:task-specific}

\smartparagraph{Pruning with domain-specific calibration sets brings clear benefits.} The proposed approach shows an overall improvement over the original Wanda paper's results. Notably, the C4 dataset does not yield the best outcomes for any task across all models and domains assessed. In each case, the top-performing sub-models are those pruned using domain-specific calibration datasets. Table~\ref{table:best_submodels} summarizes the best results achieved for the selected models\footnote{Due to space limitations, we cannot report detailed results for all the calibration sets. We will release all the code and the results upon acceptance.}, showing the sub\nobreakdash-model with the highest accuracy and the calibration dataset used. Although the sub-models generally exhibit lower accuracy than the original models, this is an expected trade-off to increase computational efficiency. Nonetheless, there is a consistent pattern across all models: the highest accuracy is obtained with calibration datasets tailored to each task, confirming our intuition that domain-specific datasets are essential for the creation of accurate, domain-specific sub-models.

\begin{figure*}[t]
    \centering
    \begin{subfigure}{.24\linewidth}
        \includegraphics[width=\textwidth]{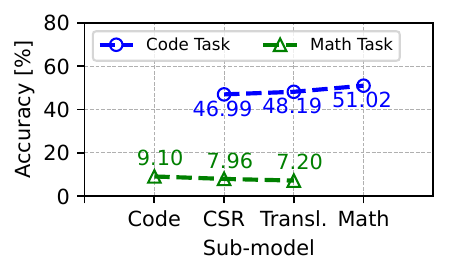}
        \caption{CodeLlama-7B.}
    \end{subfigure}
    \begin{subfigure}{.24\linewidth}
        \includegraphics[width=\textwidth]{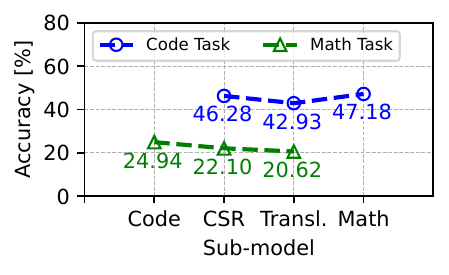}
        \caption{Mistral-7B.}
    \end{subfigure}
    \begin{subfigure}{.24\linewidth}
        \includegraphics[width=\textwidth]{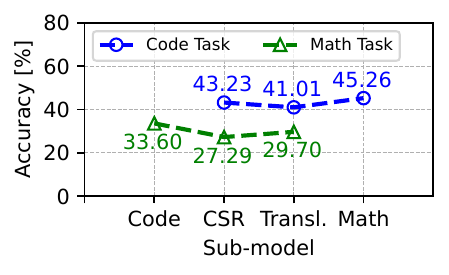}
        \caption{Gemma-7B.}
    \end{subfigure}
    \begin{subfigure}{.24\linewidth}
        \includegraphics[width=\textwidth]{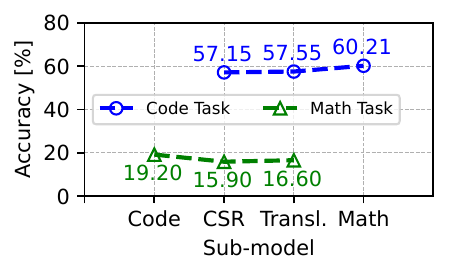}
        \caption{CodeLlama-13B.}
    \end{subfigure}
    \caption{Comparison of code-task accuracy for non-coding sub-models and of math-task accuracy for non-math sub-models.}
    \label{fig:comp}
\end{figure*}

\begin{table}[t]
\caption{Best performing sub-models.}
\vspace{-.2in}
\begin{subtable}{\linewidth}\centering
{
    \caption{CodeLlama-7B.}
    \vspace{-.05in}
    \begin{tabular}{c c c c c} 
        \multicolumn{1}{c}{\textbf{Task}} & \multicolumn{1}{c}{\textbf{Best}} & \multicolumn{1}{c}{\textbf{Pre-Prune}} & \multicolumn{1}{c}{\textbf{Post-Prune}} & \multicolumn{1}{c}{\textbf{Accuracy}} \\ 
        & \multicolumn{1}{c}{\textbf{Calib. Dataset}} & \multicolumn{1}{c}{\textbf{Accuracy}} & \multicolumn{1}{c}{\textbf{Accuracy}} & \multicolumn{1}{c}{\textbf{Drop}} \\ 
        \hline
        Code & py\_alpaca & 60.50\% & 53.03\% & 7.47\% \\
        CSR & hellaswag & 52.30\% & 49.30\% & 3.00\% \\
        Transl & wmt14 & 47.48\% & 44.40\% & 3.08\% \\
        Math & math\_inst & 17.81\% & 11.00\% & 6.81\%
    \end{tabular}
    \vspace{.09in}
}
\end{subtable}
\begin{subtable}{\linewidth}\centering
{
    \caption{Mistral-7B.}
    \vspace{-.05in}
    \begin{tabular}{c c c c c} 
        \multicolumn{1}{c}{\textbf{Task}} & \multicolumn{1}{c}{\textbf{Best}} & \multicolumn{1}{c}{\textbf{Pre-Prune}} & \multicolumn{1}{c}{\textbf{Post-Prune}} & \multicolumn{1}{c}{\textbf{Accuracy}} \\ 
        & \multicolumn{1}{c}{\textbf{Calib. Dataset}} & \multicolumn{1}{c}{\textbf{Accuracy}} & \multicolumn{1}{c}{\textbf{Accuracy}} & \multicolumn{1}{c}{\textbf{Drop}} \\ 
        \hline
        Code & py\_alpaca & 60.50\% & 47.28\% & 13.22\% \\
        CSR & ob\_qa & 61.40\% & 58.20\% & 3.20\%  \\
        Transl & wmt14 & 52.95\% & 51.31\% & 1.64\%  \\
        Math & math\_qa & 32.90\% & 27.82\% & 5.08\%
    \end{tabular}
    \vspace{.09in}
}
\end{subtable}
\begin{subtable}{\linewidth}\centering
{
    \caption{Gemma-7B.}
    \vspace{-.05in}
    \begin{tabular}{c c c c c} 
        \multicolumn{1}{c}{\textbf{Task}} & \multicolumn{1}{c}{\textbf{Best}} & \multicolumn{1}{c}{\textbf{Pre-Prune}} & \multicolumn{1}{c}{\textbf{Post-Prune}} & \multicolumn{1}{c}{\textbf{Accuracy}} \\ 
        & \multicolumn{1}{c}{\textbf{Calib. Dataset}} & \multicolumn{1}{c}{\textbf{Accuracy}} & \multicolumn{1}{c}{\textbf{Accuracy}} & \multicolumn{1}{c}{\textbf{Drop}} \\ 
        \hline
        Code & apps & 52.90\% & 47.07\% & 5.83\% \\
        CSR & ob\_qa & 40.57\% & 38.90\% & 1.67\%  \\
        Transl & o\_books & 50.41\% & 48.60\% & 1.81\%  \\
        Math & math\_o & 46.40\% & 36.16\% & 10.24\%
    \end{tabular}
    \vspace{.09in}
}
\end{subtable}
\begin{subtable}{\linewidth}\centering
{
    \caption{CodeLlama-13B.}
    \vspace{-.05in}
    \begin{tabular}{c c c c c} 
        \multicolumn{1}{c}{\textbf{Task}} & \multicolumn{1}{c}{\textbf{Best}} & \multicolumn{1}{c}{\textbf{Pre-Prune}} & \multicolumn{1}{c}{\textbf{Post-Prune}} & \multicolumn{1}{c}{\textbf{Accuracy}} \\ 
        & \multicolumn{1}{c}{\textbf{Calib. Dataset}} & \multicolumn{1}{c}{\textbf{Accuracy}} & \multicolumn{1}{c}{\textbf{Accuracy}} & \multicolumn{1}{c}{\textbf{Drop}} \\ 
        \hline
        Code & py\_alpaca & 71.00\% & 61.00\% & 10.00\% \\
        CSR & hellaswag & 54.70\% & 52.50\% & 2.20\% \\
        Transl & wmt14 & 51.47\% & 49.50\% & 1.97\% \\
        Math & math\_qa & 26.76\% & 21.68\% & 5.08\%
    \end{tabular}
}
\end{subtable}
\label{table:best_submodels}
\end{table}

\smartparagraph{Robust training yields strong post-pruning results.} Looking at the 7B models results, an interesting finding emerges in the code generation task: the Gemma model shows a relatively smaller accuracy drop after pruning compared to code-specific models, \ie CodeLlama. Although CodeLlama, specifically designed for coding tasks, was expected to preserve most of its original performance, it is actually Gemma that retains a better accuracy, with a drop of \raisebox{0.5ex}{\texttildelow}5.83\%, compared to a \raisebox{0.5ex}{\texttildelow}7.47\% drop for CodeLlama. We speculate that this result may be due to Gemma's greater resilience to pruning, likely a result of its broader training data. Unlike CodeLlama, which is primarily fine-tuned on coding data~\cite{codellama_paper}, Gemma has been exposed to a diverse range of sources, including web documents, code, and math~\cite{gemma}. This wider training data could have helped Gemma develop a deeper understanding of language, enhancing its adaptability in the pruning process. This finding is consistent with previous research~\cite{training} suggesting that a wider pre-training knowledge base contributes to better model's robustness.

\smartparagraph{Pruning retains strong CSR for user query processing.} Unlike other domains, the CSR task exhibits relatively consistent accuracy levels across different calibration datasets, even those from unrelated domains. This finding suggests that the sub-models may not gain significant benefits from the usage of CSR-specific calibration samples during pruning. The consistent accuracies across sub-models indicate that they possess a robust understanding of common knowledge and everyday reasoning principles, likely due to their training on web documents or datasets rich in general knowledge. Therefore, additional calibration with CSR-specific samples during pruning appears to offer limited benefits. Since each model requires a strong base of common knowledge to effectively process user queries, it is likely that essential weights for CSR are consistently retained across all sub-models.

\smartparagraph{Task complexity influences post-pruning accuracy.} In the language translation task, we observe a similar trend to that of code generation: translation-specific sub-models retain higher accuracy, confirming the effectiveness of our approach. Interestingly, translation sub-models experience a much smaller accuracy drop ($\le$3\%) than those focused on code generation tasks ($\ge$5\%). We hypothesize that this notable difference arises from the fundamental nature of the tasks. Translation typically involves a more direct mapping between input and output sequences across languages, while code generation requires transforming natural language into executable code, which demands a deep understanding of programming logic, syntax, and semantics. Consequently, code-generation sub-models may lose essential information during pruning, leading to the observed larger accuracy drop. This observation also suggests that resulting sub-models could be utilized for performing specific sub-tasks within a domain (\eg line-by-line code translation between programming languages) rather than broader tasks (\eg generating complete code blocks).

\begin{table}[b]
\vspace{-.1in}
    \caption{Language-specific sub-models evaluation.}
\centering
\resizebox{\linewidth}{!}{%
    \begin{tabular}{c c *{4}{d{3.3}} }
        & & \multicolumn{4}{c}{\textbf{Evaluation Datasets}} \\
        \cmidrule(lr){3-6}
        \multicolumn{1}{c}{\textbf{Language}} & \multicolumn{1}{c}{\textbf{Calib. Dataset}} & \multicolumn{1}{c}{\textbf{HE (Python)}} & \multicolumn{1}{c}{\textbf{HE (Java)}} & \multicolumn{1}{c}{\textbf{HE (C++)}} & \multicolumn{1}{c}{\textbf{HE (JS)}} \\ 
        \hline
        - & None &  71.00\%&  68.3\%&  64.1\%&  67.7\%\\
        Python & py\_alpaca&  \multicolumn{1}{c}{\bfseries 61.00\%}&  55.2\%&  54.2\% &  59.31\%\\
        Java & code-java &  56.1\%&  \multicolumn{1}{c}{\bfseries 60.1\%}&  54.2\%&  55.6\%\\
        C++ & cpp\_dataset &  56.01\%&  54.2\%&  \multicolumn{1}{c}{\bfseries 57.1\%}&  56.3\%\\
        Javascript & code-javascript &  56.7\%&  55.7\%&  50.9\%&  \multicolumn{1}{c}{\bfseries 60.2\%}
    \end{tabular}
    }
    \label{table:language-specific-res}
\end{table}

\begin{figure*}[t]
    \centering
    \begin{subfigure}{\linewidth}
    \centering
        \includegraphics[width=.6\columnwidth]{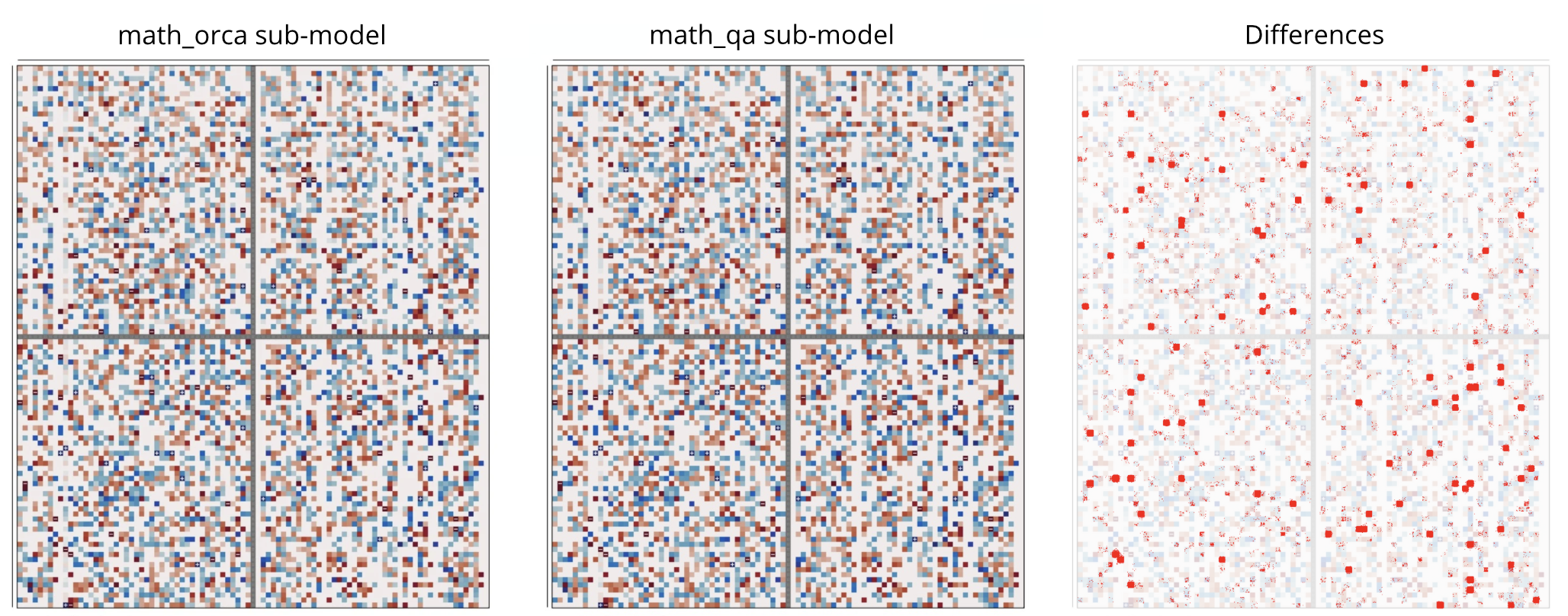}
        \caption{Math vs.\ Math. The differences between the two sub-models' weights are minimal.}
        \label{fig:math-submodels-penzai}
    \end{subfigure}
    \begin{subfigure}{\linewidth}
        \centering
        \includegraphics[width=.6\columnwidth]{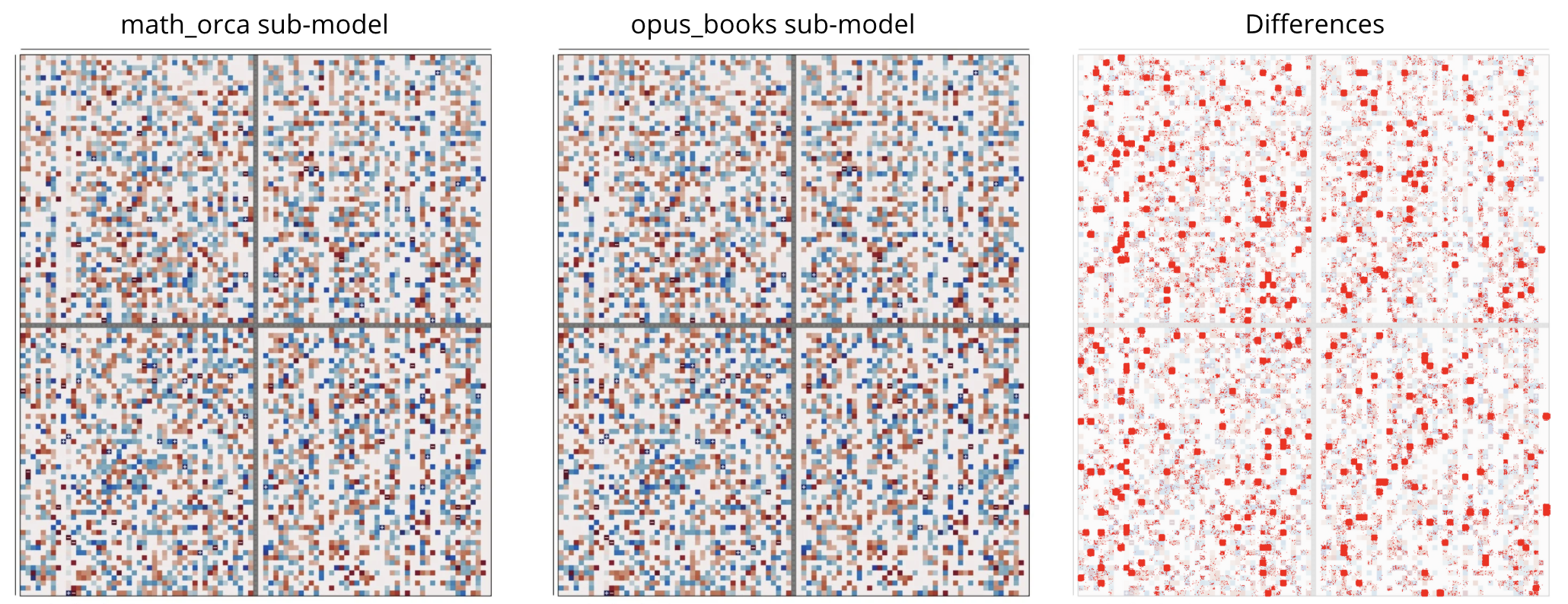}
        \caption{Math vs.\ Translation. A good portion of the weights differs.}
        \label{fig:math-tr-submodels-penzai}
    \end{subfigure}
    \caption{The first two plots (left and center) represent the visualization of the \texttt{q\_proj} weight matrices of layer 12 of two sub-models obtained by pruning Gemma on two different datasets. The last plot (right) represents the weights that differ between the two matrices (red dots).}
    \label{fig:submodels-penzai}
    \vspace{-.2in}
\end{figure*}

\begin{figure}[b]
    \vspace{-.1in}
    \centering
    \begin{subfigure}{.45\linewidth}
        \includegraphics[width=\textwidth]{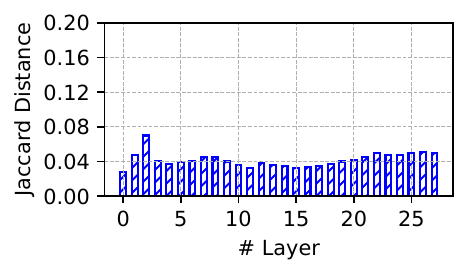}
        \caption{}
        \label{fig:mathmath-jaccard}
    \end{subfigure}
    \begin{subfigure}{.45\linewidth}
        \includegraphics[width=\textwidth]{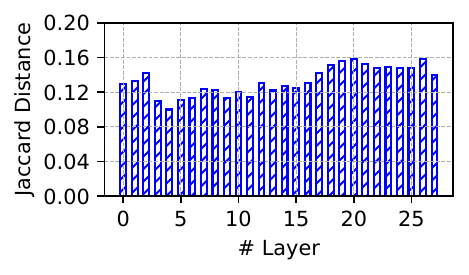}
        \caption{}
        \label{fig:math-tr-jaccard}
    \end{subfigure}
    \caption{Jaccard distance of Gemma pruned on (a) math\_qa (math) and math\_orca (math), and (b) math\_qa (math) and opus\_books (translation).}
    \label{fig:jaccard}
    \vspace{-.2in}
\end{figure}

\smartparagraph{Logical reasoning skills retain cross-task accuracy.} We observe that code and math sub-models experience only a slight drop in accuracy when evaluated on tasks from the other domain (\ie math models on code tasks and vice versa). Fig.~\ref{fig:comp} shows the accuracy of code (blue line) and math (green line) tasks conducted with non-code and non-math sub-models for the selected 7B models. It is evident that code-specific sub-models perform slightly better on math tasks, and the same applies for math-specific sub-models on coding tasks. This interesting behavior can be explained by the shared skills and similarities between coding and mathematical reasoning. Both areas require logical thinking, problem-solving skills, and a solid  understanding of numerical concepts. As a result, sub-models specialized in one domain may retain sub-areas of knowledge relevant to the other, enabling them to achieve higher accuracy levels.

\smartparagraph{Proper calibration sets extract programming-language-specific sub-models.} After evaluating the impact of task-specific calibration sets in pruning, which supports our approach for extracting domain-specific sub-models effectively, we explore whether similar outcomes could be achieved with programming-language-specific sub-models. In previous experiments, we solely focused on Python in the code generation task. Here, we expand our approach to extract three additional sub-models: Java, C++, and JavaScript. We select \mbox{CodeLlama-13B} as the base model, given the higher accuracy of its sub-models. Table~\ref{table:language-specific-res} presents the performance of each sub-model pruned with different programming language datasets as calibration sets. The evaluation spans four distinct versions of HumanEval, each tailored to a specific programming language. A significant observation is the strong language specificity of each sub-model: they achieve high accuracy within their own language but show an accuracy drop when applied to other languages. This finding demonstrates that it is possible to effectively extract programming-language-specific sub-models from foundational models with suitable calibration sets, and that such sub-models retain acceptable accuracy w.r.t. base models.

\subsection{Sub-Models Structure Comparison}\label{ss:submodel-structure}

In order to determine whether our approach is effectively creating domain-specific sub-models, we conducted a comparison of the parameter masks of each layer of the \mbox{pruned models}.

\smartparagraph{Distinct weight patterns emerge in \mbox{domain-specific} \mbox{sub-models}.} We analyzed the internal structure of the obtained domain-specific sub-models to determine whether the pruning method influences their specificity to certain domains. Due to space limitations, we include results only for Gemma-7B, which displays the most significant differences and has fewer layers, providing a clearer visualization. Fig.~\ref{fig:submodels-penzai} presents visual representations of the post-pruning weights of the \texttt{q\_proj} matrix\footnote{\texttt{q\_proj} refers to the Query projection matrix in the attention mechanism.} in the 12th layer of the model, generated using the \textit{penzai} tool~\cite{penzai}. Specifically, Fig.~\ref{fig:math-submodels-penzai} compares two math-specific sub-models, while Fig.~\ref{fig:math-tr-submodels-penzai} contrasts one math-specific sub-model with a translation-specific sub-model. In each figure, the left and center plots report weights as colored squares, where darker squares indicate higher values of the corresponding weights, while the right-most plot highlights the differences in weights between the two matrices (red dots). This visualization allows us to qualitatively assess that sub-models from distinct domains exhibit notable disparities compared to those within the same domain, as the number of red dots in Fig.~\ref{fig:math-submodels-penzai} (math vs.\ math) is markedly lower than the ones in Fig.~\ref{fig:math-tr-submodels-penzai} (math vs.\ translation).

\smartparagraph{Quantitative analysis of sub-models weight patterns.} We deepen the understanding of weight differences through an analytical per-layer comparison using the Jaccard distance. Fig.~\ref{fig:jaccard} depicts the per-layer Jaccard distance of the same Gemma sub-models shown in Fig.~\ref{fig:submodels-penzai}. Specifically, Fig.~\ref{fig:mathmath-jaccard} reports the comparison between sub-models within the same domain (math), while Fig.~\ref{fig:math-tr-jaccard} compares sub-models from different domains (math and translation). The clear contrast indicates that sub-models within the same domain tend to have more similar weight distributions, while those from different domains show more divergent weight distributions. These results suggest the existence of task-specific weights essential for particular domains, while other weights may be less relevant within those domains. However, these less relevant weights in one domain could be critical for others, explaining the significant differences observed between models from different domains. 

\vspace{.05in}
In summary, through our qualitative and quantitative analyses, we demonstrate that domain-specific tasks activate different regions within LLMs, resulting in distinct parameter masks, thereby extending and confirming similar findings from previous studies~\cite{dpruner} but focusing on resource-efficient unstructured pruning. Therefore, we find that domain-specific sub-models can be effectively extracted from foundational LLMs using suitable calibration datasets.


\section{Conclusions}
In this study, we demonstrate that domain-specific sub-models can be efficiently extracted from foundational LLMs using unstructured pruning (\ie Wanda). This approach was applied across three distinct domains: math, CSR, and language translation. Our findings reveal that the resulting sub-models maintain acceptable accuracy when compared to their full-model counterparts, and that sub-models derived using domain-specific calibration sets outperform those created with unrelated data, highlighting the importance of selecting appropriate calibration datasets. We are the first ones to extract four programming-language-specific sub-models, further confirming our findings. Furthermore, we provide the first analytical evidence that domain-specific tasks activate unique LLMs regions with unstructured pruning, indicating that certain weights are essential for maintaining domain-specific performance. 

Future work could extend these experiments by incorporating additional models of different sizes and exploring a wider range of programming languages. Additionally, we also plan to develop a novel inference system that utilizes the extracted sub-models and dynamically selects the most appropriate according to user requirements (similarly to Composition of Experts~\cite{samba}).



\section*{Acknowledgments}

We would like to thank the anonymous reviewers for their insightful comments and suggestions on this paper. This work has been partially supported by Knut and Alice Wallenberg Foundation (Wallenberg Scholar Grant for Prof. Dejan Kosti\'c), Vinnova (the Sweden's Innovation Agency), the Swedish Research Council (agreement No. 2021-04212), and KTH Digital Futures. We acknowledge EuroHPC Joint Undertaking for awarding us access to the Leonardo supercomputer located at the CINECA data center in Bologna, Italy.

\bibliographystyle{IEEEtran}
\begin{small}
\bibliography{bibliography}
\end{small}

\end{document}